\documentclass{article}
\usepackage[preprint,nonatbib]{neurips_2019}

\usepackage[utf8]{inputenc} 
\usepackage[T1]{fontenc}    
\usepackage{hyperref}       
\usepackage{url}            
\usepackage{booktabs}       
\usepackage{amsfonts}       
\usepackage{nicefrac}       
\usepackage{microtype}      

\usepackage{amsmath}
\usepackage{algorithm, algorithmic}
\usepackage{amsfonts, amssymb}
\usepackage{multirow}
\usepackage{mathtools}
\usepackage{arydshln}
\usepackage{refcount}
\usepackage{listings}
\usepackage{hyperref}

\DeclarePairedDelimiter{\ceil}{\lceil}{\rceil}

\newtheorem{theorem}{Theorem}[section]
\newtheorem{corollary}[theorem]{Corollary}
\newtheorem{lemma}[theorem]{Lemma}

\newtheorem{assumption}[theorem]{Assumption}

\renewcommand\thefigure{\arabic{figure}}
\renewcommand\thetable{\arabic{table}}

\def\BibTeX{{\rm B\kern-.05em{\sc i\kern-.025em b}\kern-.08em
    T\kern-.1667em\lower.7ex\hbox{E}\kern-.125emX}}

\title{Residual Networks Behave Like Boosting Algorithms\thanks{This work was supported by and completed whilst author was at Suncorp Group Limited.}}

\author{%
  Chapman Siu \\
  Faculty of Engineering and Information Technology \\
  University of Technology Sydney, Australia \\
  \texttt{chpmn.siu@gmail.com} \\
}

\begin{document}

\maketitle

\begin{abstract}

We show that Residual Networks (ResNet) is equivalent to boosting feature representation, without any modification to the underlying ResNet training algorithm. A regret bound based on Online Gradient Boosting theory is proved and suggests that ResNet could achieve Online Gradient Boosting regret bounds through neural network architectural changes with the addition of a shrinkage parameter in the \emph{identity skip-connections} and using residual modules with max-norm bounds. Through this relation between ResNet and Online Boosting, novel feature representation boosting algorithms can be constructed based on altering residual modules. We demonstrate this through proposing decision tree residual modules to construct a new boosted decision tree algorithm and demonstrating generalization error bounds for both approaches; relaxing constraints within BoostResNet algorithm to allow it to be trained in an out-of-core manner. We evaluate convolution ResNet with and without shrinkage modifications to demonstrate its efficacy, and demonstrate that our online boosted decision tree algorithm is comparable to state-of-the-art \emph{offline} boosted decision tree algorithms without the drawback of offline approaches.

\end{abstract}

\section{Introduction}

Residual Networks (ResNet) \cite{He2016DeepRL} have previously had a lot of attention due to performance, and ability to construct ``deep'' networks while largely avoiding the problem of vanishing (or exploding) gradients. Some attempts have been made in explaining ResNets through: unravelling their representation\cite{Veit2016RNB};  observing identity loops and finding no spurious local optima\cite{hardt2016identity}; and reinterpreting residual modules as weak classifiers which allows sequential training under boosting theory\cite{HuangALS17}.


Empirical evidence shows that these deep residual networks, and subsequent architectures with the same parameterizations, are easier to optimize. They also out-perform non-residual ones, and have consistently achieved state-of-the-art performance on various computer vision tasks such as CIFAR-10 and ImageNet\cite{He2016DeepRL}.



\subsection{Summary of Results}

We demonstrate the equivalence of ResNet and Online Boosting in Section \ref{sec:equal}. We show that the layer by layer boosting method of ResNet has an equivalent representation of additive modelling approaches first demonstrated in Logitboost \cite{friedman2000additive} which boost feature representations rather than the label directly, i.e. ResNet can be framed as an Online Boosting algorithm with composite loss functions. Although traditional boosting results may not apply as ResNet are not a naive weighted ensemble, we can refer to them as Online Boosting analogues which presents regret bound guarantees. We demonstrate that under ``nice'' conditions for the composite loss function, the regret bound for Online Boosting holds, and by extension also applies for ResNet architectures. 

Taking inspiration from Online Boosting, we also modify the architecture of ResNet with an additional \emph{learnable} shrinkage parameter (vanilla ResNet can be interpreted as Online Boosting algorithm where the shrinkage factor is fixed/unlearnable and set to $1$). As this approach only modifies the neural network architecture, the same underlying ResNet algorithms can still be used. 


Experimentally, we compare vanilla ResNet with our modified ResNet using \emph{convolutional neural network residual network} (ResNet-CNN) on multiple image datasets. Our modified ResNet shows some improvement over vanilla ResNet architecture. 

We also compare our boosted decision tree \emph{neural decision tree residual network} on multiple benchmark datasets and their results against other decision tree ensemble methods, including \emph{Deep Neural Decision Forests} \cite{dndf}, neural decision trees ensembled via \emph{AdaNet} \cite{cortes17a}, and off-the-shelf algorithms (gradient boosting decision tree/random forest) using \emph{LightGBM} \cite{lightgbm}. In our experiments, \emph{neural decision tree residual network} showed superior performance to neural decision tree variants, and comparable performance to offline tradition gradient boosting decision tree models. 

\subsection{Related Works}

In recent years researchers have sought to understand why ResNet perform the way that they do. The BoostResNet algorithm reinterprets ResNet as a \emph{multi-channel telescoping sum boosting} problem for the purpose of introducing a new algorithm for sequential training \cite{HuangALS17}, providing theoretical justification for the representational power of ResNet under linear neural network constraints\cite{hardt2016identity}. One interpretation of residual networks is as a collection of many paths of differing lengths which behave like a shallow ensemble; empirical studies demonstrate that residual networks introduce short paths which can carry gradients throughout the extent of very deep networks \cite{Veit2016RNB}.


\textbf{Comparison with BoostResNet and AdaNet}

Combining Neural Networks and boosting has previously been explored in architectures such as AdaNet \cite{cortes17a} and BoostResNet \cite{HuangALS17}. We seek to understand \emph{why} ResNet achieves their level of performance, without altering how ResNet are trained.

In the case of BoostResNet, the distribution must be explicitly maintain over all examples during training and parts of ResNet are trained sequentially which cannot be updated in a truly online, out-of-core manner. And in the case of AdaNet, which do not always work for ResNet structure, additional feature vectors are sequentially added, and chooses their own structure during learning. In our proposed approach, we do not require these modifications, and can train the model in the same way as an unmodified ResNet. A ResNet style architecture is a special case of AdaNet, so AdaNet generalization guarantee applies here and our generalization analysis is built upon their work. Furthermore we also demonstrate Neural Decision Trees belong to same family of feedforward neural networks as AdaNet, so AdaNet generalization guarentee also applies to Neural Decision Tree ResNet modules and our generalization analysis is built upon their work.



\section{Preliminaries}

In this section we cover the background of \emph{residual neural networks} and \emph{boosting}. We also explore the conditions which enable regret bounds in Online Gradient Boosting setting and the class of feedforward neural networks for AdaNet generalization bounds.

\subsection{Residual Neural Networks}

A \emph{residual neural network} (ResNet) is composed of stacked entities referred to as residual blocks. 

\textbf{A Residual Block of ResNet} contains a module and an identity loop. Let each module map its input $x$ to $f_t(x)$ where $t$ denotes the level of the module, and where $f_t(x)$ is typically a sequence of convolutions, batch normalizations or non-linearities. These formulations may differ depending on context and the model architecture. 

We denote the output of the $t$-th residual block to be $g_{t+1}(x)$

\begin{equation}\label{eq:1}
    g_{t+1}(x) = f_t(g_t(x)) + g_t(x)
\end{equation}

where $x$ is the input of the ResNet. 

\label{sec:outputresnet}
\textbf{Output of ResNet} has a recursive relation specified in equation \ref{eq:1}, then output of the $T$-th residual block is equal to the summation of lower module outputs, i.e., $g_{T+1}(x) = \sum_{t=0}^T f_t(g_t(x))$, where $g_0(x) = 0$ and $f_0(g_0(x)) = x$. For classification tasks, the output of a ResNet is rendered after a linear classifier $\mathbf{w} \in \mathbb{R}^{n \times C}$ on representation $g_{T+1}(x)$ where $C$ is the number of classes, and $n$ is the number of channels: 

\begin{equation}\label{eq:2}
    \hat{y} = \tilde{\sigma}(\mathbf{w}^\top g_{T+1}(x)) = \tilde{\sigma}\left(\sum_{t=0}^T \mathbf{w}^\top f_t(g_t(x))\right)
\end{equation}

where $\sigma(\cdot)$ denotes a map from classifier output to labels. For example, $\sigma$ could represent a softmax function. 

\subsection{Boosting}

The goal of boosting is to combine weaker learners into a strong learner. There are many variations to boosting. For example, in AdaBoost and its derivatives, we require the boosting algorithm to choose training sets for the weak classifier to force it to make novel inferences \cite{friedman2000additive}. This was the approached used by BoostResNet \cite{HuangALS17}. In gradient boosting, this requirement is removed through training against pseudo-residual and can even be extended to the online learning setting \cite{beygelzimer2015online}.

In either scenario, boosting can be viewed as an additive model or linear combinations of $T$ models $\tilde{F}(x) = \sum_{t=1}^T \alpha_t h_t(x)$, where $h_t(x)$ is a function of the input $x$ and $\alpha_t$ is the corresponding multiplier for the $t$-th model \cite{friedman2000additive}\cite{beygelzimer2015online}. 

\textbf{LogitBoost} is an algorithm first introduced in ``Additive Logistic Regression: A Statistical View of Boosting''\cite{friedman2000additive}, which introduces boosting on the input feature representation, including neural networks. In the general binary classification scenario, the formulation relies on boosting over the \emph{logit} or softmax transformation

\begin{equation}\label{eq:3}
    \hat{y} = \frac{e^{\tilde{F}(x)}}{1+e^{\tilde{F}(x)}}=\tilde{\sigma}(\tilde{F}(x))=\tilde{\sigma}\left(\sum_{t=1}^T \alpha_t h_t(x)\right)
\end{equation}

Where $\tilde{\sigma}$ represents the softmax function. This form is similar to the linear classifier layer which is used by ResNet algorithm. 

\textbf{Online Boosting} introduces a framework for training boosted models in an online manner. Within this formulation, there are two adjustments which are required to make offline boosting models online. First, the partial sums $\hat{y}^{i-1}$ (where $\hat{y}^i$ represents the predictions of the $i$-th model) is multiplied by a \emph{shrinkage} factor, which is tuned using gradient descent. Second, the partial sums $\hat{y}^i$ outputs are to be bounded \cite{beygelzimer2015online}. 

The bounds presented for online gradient boosting are based on regret. The regret $R_\mathcal{A}(T)$ of a learner is defined as the difference between the total loss from the learner and the total learner of the best hypothesis in hindsight

\begin{align*}
R_{\mathcal{A}}(T) = \sum_{t=1}^T \ell_t(h_t(x_t)) - \min_{h^* \in \mathcal{F}} \sum_{t=1}^T \ell(f^*(x_t))
\end{align*}

Online gradient boosting regret bounds applies can be applied to any linear combination of a give base weak learner with a convex, linear loss function that is Lipschitz constant bounded by $1$. 

\begin{corollary}\label{th_be}
(From Corollary 1 \cite{beygelzimer2015online}) Let the learning rate $\eta \in [\frac{1}{N}, 1]$, number of weak learners $N$, be given parameters. Algorithm \ref{alg} is an online learning algorithm for $span(\mathcal{F})$ for set of convex, linear loss functions with Lipschitz constant bounded by $1$ with the following regret bound for any $f \in span(\mathcal{F})$:

\begin{align*}
R^\prime_f(T) \leq (1&-\frac{\eta}{||f||_1})^N \Delta_0 + \\
&O(||f||_1 \cdot(\eta T + R(T) + \sqrt{T}))
\end{align*} 
where $\Delta_0 := \sum_{t=1}^T \ell^\psi_t(0) - \ell^\psi_t(f(x_t))$, or the initial error, and $R(T)$ is  the regret or excess loss for the base learner algorithm.
\end{corollary}

The regret bound in this theorem depends on several conditions; the requirement that for any weak learner $\mathcal{A}$, that it has a finite upper bound, i.e. $||\mathcal{A}(x)|| \leq D$, for some $D$, and the set of loss functions constraints an efficiently computable subgradient $\nabla \ell (y)$ has a finite upper bound.

Compared with boosting approach used in BoostResNet which is based on AdaBoost\cite{HuangALS17}, the usage of the online gradient boosting algorithm does not require maintaining an explicit distribution of weights over the whole training data set and is a ``true'' online, out-of-core algorithm. Leveraging online gradient boosting allows us to overcome the constraints of BoostResNet approach.

\textbf{AdaNet Generalization Bounds} for feedforward neural networks defined to be a multi-layer architecture where units in each layer are only connected to those in the layer below has been provided by \cite{cortes17a}. It requires the weights of each layer to be bounded by $l_p$-norm, with $p \geq 1$, and all activation functions between each layer to be coordinate-wise and $1$-Lipschitz activation functions. This yields the following generalization error bounds provided by Lemma 2 from \cite{cortes17a}: 

\begin{corollary} \label{th_gen}
(From Lemma 2 \cite{cortes17a}) Let $\mathcal{D}$ be distribution over $\mathcal{X}\times \mathcal{Y}$ and $S$ be a sample of $m$ examples chosen independently at a random according to $\mathcal{D}$. With probability at least $1-\delta$, for $\theta > 0$, the strong decision tree classifier $F(x)$ satisfies that 
$$
R(f) \leq \hat{R}_{S, \rho}(f) + \frac{4}{\rho}\sum_{k=1}^l \lvert w_k\rvert_1 \mathcal{R}_m(\tilde{\mathcal{H}}_k) + \frac{2}{\rho}\sqrt{\frac{\log l}{m}} $$
 $$ + C(\rho, l, m, \delta) $$
 $$\text{where } C(\rho, l, m, \delta) = \sqrt{\ceil{\frac{4}{\rho^2} \log (\frac{\rho^2m}{\log l})} \frac{\log l}{m} + \frac{\log \frac{2}{\delta}}{2m}} $$
\end{corollary}

As this bound depends only on the logarithmically on the depth for the network $l$ this demonstrates the importance of strong performance in the earlier layers of the feedforward network.

Now that we have a formulation for ResNet and boosting, we explore further properties of ResNet, and how we may evolve and create more novel architectures. 

\section{ResNet are Equivalent to a Boosted Model}
\label{sec:equal}

As we recall from equations \ref{eq:2} and \ref{eq:3}, ResNet indeed have a similar form to LogitBoost. In this scenario, both formulations aim to boost the underlying feature representation. One consequence of the ResNet formulation is that the linear classifier $\mathbf{w}$, would be a shared linear classifier across all all ResNet modules. 





\begin{assumption}\label{assum}
The $t$-th residual module with a trainable linear classifier layer $\mathbf{w}$ defined by

\begin{equation}
    \tilde{h}_t(x) := \mathbf{w}^\top f_t(g_t(x))
\end{equation}

Is a weak learner for all $t \geq 0$. We will call this weak learner the hypothesis module.

\end{assumption}

This assumption is required to ensure that $\tilde{h}_t(x)$ is a weak learner to adhere to learning bounds proposed in Corollary \ref{th_be}. 
We show that different ResNet modules variants used in our experiments assumption in Sections \ref{cnn_exp}.

Overall this demonstrates that the proposed framework is equivalent to traditional boosting frameworks which boost on the feature representation. However, to further analyse the algorithmic results, we need to first consider additional restrictions which are placed within the ``Online Boosting Algorithm'' framework. 

\subsection{Online Boosting Considerations}

Our representation is a special case of online gradient boosting as shown in Algorithm \ref{alg}, our regret bound analysis is built upon work in \cite{beygelzimer2015online}. The regret bounds for an online boosting algorithm that competes with linear combination of the base weak learner applies when used for a class of  convex, linear loss function with Lipschitz constant bounded by $1$.

\begin{algorithm}
\caption{Online Boosting for Composite Loss Functions for $\text{span}(\mathcal{F})$}
\label{alg}
\begin{algorithmic}[1]
\STATE Maintain $N$ copies of the algorithm $\mathcal{A}$, denoted $\mathcal{A}^1, \mathcal{A}^2, \dots, \mathcal{A}^N$ and choose step size parameter $\eta \in [\frac{1}{N}, 1]$
\STATE For each $i$, initialize $\theta^i = 0$. 
\FOR{$t=1$ to $T$}
\STATE Receive example $\mathbf{x}_t$
\STATE Define $\mathbf{F}^0_t = 0$
\FOR{$i=1$ to $N$}
\STATE $\mathbf{F}_t^i = (1-\theta^i )\mathbf{F}_t^{i-1} + \eta\mathcal{A}^i(\mathbf{x}_t)$, where $(1-\theta^i )$ is our shrinkage factor for algorithm $\mathcal{A}^i$
\ENDFOR
\STATE Predict $\mathbf{y}_t = \psi^{-1}(\mathbf{F}^N_t)$
\STATE Obtain loss function $\ell_t^\psi$ and the model suffers loss $\ell^\psi_t(\mathbf{F}_t)$, which is equivalent to equivalently $\ell_t(\mathbf{y}_t)$ 
\FOR{$i=1$ to $N$}
\STATE Pass loss based on partial sums of $\mathbf{F}_t^{i-1}$ to $\mathcal{A}^i$, i.e. in descent direction $\nabla \ell_t^\psi(\mathbf{F}_{t}^{i-1}) \cdot \mathbf{y}$
\STATE Update $\theta^i \in [0, \eta]$ using online gradient descent
\ENDFOR
\ENDFOR
\end{algorithmic}
\end{algorithm}

This algorithm yields the regret bound for algorithm \ref{alg}, which is directly from Corollary 1 from \cite{beygelzimer2015online} in Corollary \ref{th_be}. We will provide further analysis of this algorithm; in particular the validity of composite loss functions in Section \ref{sec:analysis}.

\section{Analysis of Online Boosting for Composite Loss Functions}\label{sec:analysis}

In this section we provide analysis on corollary \ref{th_be}. Corollary \ref{th_be} holds for the learning algorithm with losses in $\mathcal{C}$, where $\mathcal{C}$ is defined to be set of convex, linear loss functions with Lipschitz constant bounded by $1$. Next we describe the conditions in which composite loss functions $\ell^\psi$ belongs in $\mathcal{C}$. 

A composite loss function $\ell^\psi$, where $\psi$ is the link function, belongs to $\mathcal{C}$ if $\psi$ is the canonical link function. This has been shown to be a sufficient but not necessary condition for canonical link to lead to convex composite loss functions \cite{reid2010composite}.

\begin{lemma}
Composite loss functions retain smoothness
\end{lemma}

\textbf{Proof:} If \(\psi^{-1}\) satisfies Lipschitz continuous function (e.g.~logistic
function/softmax, as its derivative is bounded everywhere), then the
composite loss is also Lipschitz constant, as composition of functions
which are Lipschitz constant is also Lipschitz constant. As if \(f_2\)
has Lipschitz constant \(L_2\) and \(f_1\) has Lipschitz constant
\(L_1\) then

\begin{align*}
\lvert f_2(f_1(x)) - f_2(f_1(y))\rvert &\leq L_2 \lvert f(x) - f_1(y) \rvert \\
&\leq L_1 L_2 \lvert x - y \rvert
\end{align*}

Hence, if $\psi^{-1}$ has Lipschitz constant bounded by $1$, then the composition of the particular loss function with Lipschitz constant of $1$ also has Lipschitz constant bounded by $1$ and belongs to the base loss function class. An example of such a link function is the logit function, which has a Lipschitz constant of 1 and is the canonical link function for log loss (cross entropy loss), which suggests that the composite loss function is indeed convex and belongs in loss function class $\mathcal{C}$

This demonstrates ResNet which boost on the feature representation and have a logit link satisfies regret bound as shown in Collorary \ref{th_be}.

\section{Recovering Loss For Intermediary Residual Modules}

\begin{figure}[ht] 
\vskip 0.2in
\begin{center}
\centerline{\includegraphics[width=0.7\linewidth]{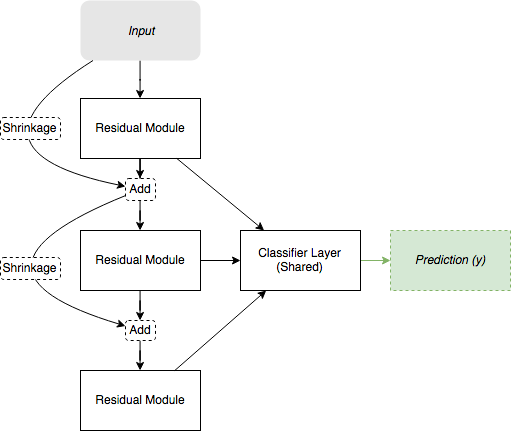}}
\caption{The architecture of a modified residual network (three residual modules) with shrinkage parameter and shared linear classifier}\label{fig_shared_arch}
\end{center}
\vskip -0.2in
\end{figure}

When analysing the ResNet and Online Boosting algorithm, the Online Boosting algorithm requires the gradient of the underlying boosting function to be recovered as part of the update process. This is shown in line 12 within Algorithm \ref{alg}. One approach to tackle this challenge was suggested in BoostResNet where a common auxiliary linear classifier is used across all residual modules, however this approach was not explored in the work as BoostResNet was focused on sequential training of residual modules, and such a constraint was deemed inappropriate. Instead BoostResNet would construct different linear classifier layers which were dropped at every stage when the residual modules have been trained.  

Our approach to remediate this is to formulate the ResNet architecture as a single-input, multi-output training problem, whereby each residual module will have an explicit `shortcut' to the output labels whilst sharing the same linear classifier layer. This architecture is shown in Figure \ref{fig_shared_arch}. 

\textbf{Remark:} It has been demonstrated that through carefully constructing a ResNet, the last layer need not be trained, and instead can be a fixed random projection\cite{hardt2016identity}. This has been demonstrated through theoretical justifications in linear neural networks.

\begin{algorithm}
\caption{Online Boosting Algorithm as ResNet with Shrinkage}\label{alg2}

\begin{algorithmic}[1]
\STATE Maintain $N$ ResNet Modules $\mathcal{\tilde{A}}$, $N$ shrinkage layers $\theta$, linear classifier layer $\mathbf{w}$  and choose step size parameter $\eta \in [\frac{1}{N}, 1]$, constructed as per Figure \ref{fig_shrink}
\STATE For each $i$, initialize shrinkage layer $\theta^i = 0$.
\STATE Define $\mathbf{F}^0_t = 0$
\FOR{$t=1$ to $T$}
\STATE \COMMENT{\textbf{Feed Forward}}
\STATE Receive example $\mathbf{x}_t$
\FOR{$i=1$ to $N$}
\STATE $\mathbf{F}_t^i = (1-\theta^i )\mathbf{F}_t^{i-1} + \eta
\mathbf{w}^\top\mathcal{\tilde{A}}^i(\mathbf{x}_t)$
\STATE Predict and output $\mathbf{y}^i_t = \psi^{-1}(\mathbf{F}^i_t)$
\ENDFOR

\STATE \COMMENT{\textbf{Back Propagation}} 
\FOR{$i=1$ to $N$}
\STATE Update all layers in the subnetwork up to ResNet module $i$ via back propagation using the final prediction output $\mathbf{y}_t^i$
\ENDFOR
\ENDFOR
\end{algorithmic}
\end{algorithm}

In our ResNet algorithm, if the linear classifier layer is shared, then the model would be framed as a single input, multi-output residual network, where the outputs are all predicting the same output $\mathbf{y}$. The predicted output of the network, which corresponds to each of the weak learners $\mathbf{w}^T\mathcal{\tilde{A}}^i$ would correspond to $y_t^i$ on lines 8 and 9 of Algorithm \ref{alg2}. Through this setup, it allows each residual module, $\mathcal{\tilde{A}}^i$ to be updated by back propagation with respect to the label $\mathbf{y}$ in the same manner as line 12 in Algorithm \ref{alg}. In a similar manner the shrinkage layers $\theta^i$ in Algorithm \ref{alg2} would be updated as shown in Algorithm \ref{alg} as per line $13$.

\begin{figure}[ht]
\vskip 0.2in
\begin{center}
\centerline{\includegraphics[width=0.3\linewidth]{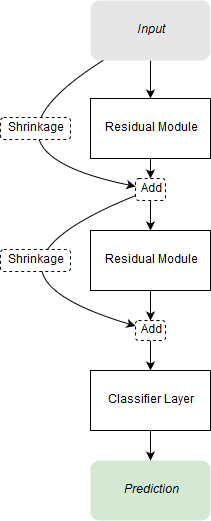}}
\caption{The architecture of a modified residual network two modules and with shrinkage layers.}\label{fig_shrink}
\end{center}
\vskip -0.2in
\end{figure}

Through unravelling a ResNet, the paths of a ResNet are distributed in a binomial manner\cite{Veit2016RNB}, that is, there is one path that passes through $n$ modules and $n$ paths that go through one module, with an average path length of $n/2$ \cite{Veit2016RNB}. This means that even without a shared layer, there will be paths within the ResNet framework where the gradient is recovered to the residual modules. This approach is shown by figure \ref{fig_shrink}, and has an identical setup as algorithm \ref{alg2} except in the back propagation step, we update all layers based on the whole network using output $\mathbf{y}_t^N$ only.

\textbf{Remark:}\label{remark} if a residual network is reframed as the Online Boosting Algorithm \ref{alg}, it would be equivalent to choosing $\eta = 1$, with $\theta^i = 0$ being fixed or untrainable. For the regret bounds to hold, we require shrinkage parameter $\theta^i$ to be trainable, and the outputs of each residual module to be bounded by a predetermined max-norm.

In Section \ref{experiments} we will provide empirical evidence validating both approaches.

\subsection{Neural Decision Tree}

Another popular application of boosting algorithms is through the construction of decision trees. In order to demonstrate how ResNet could be used to boost a variety of models with different residual module representations, we describe our construction for our Neural Decision Tree ResNet and the associated generalization error analysis.

\subsubsection{Construction of Neural Decision Tree and Generalization Error Analysis}

\begin{figure}[ht]
\vskip 0.2in
\begin{center}
\centerline{\includegraphics[width=0.5\linewidth]{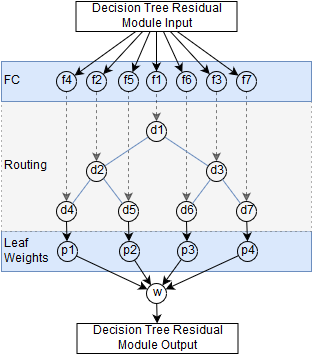}}
\caption{Decision Tree Residual Module based on Decision Tree Algorithm in ``Deep Neural Decision Forests'' \cite{dndf}}\label{fig:dt}
\end{center}
\vskip -0.2in
\end{figure}

To demonstrate decision tree formulation based on Deep Neural Decision Forests belongs to this family of neural network models, consider the residual module is shown by Figure \ref{fig:dt}, where the split functions are realized by randomized multi-layer perceptrons \cite{dndf}. This construction is a neural network has $3$ sets of layers that belongs to family of artificial neural networks defined by \cite{cortes17a}; which require the weights of each layer to be bounded by $l_p$-norm, with $p \geq 1$, and all activation functions between each layer to be coordinate-wise and $1$-Lipschitz activation functions. The size of these layers are based on a predetermined number of nodes $n$ with a corresponding number of leaves $\ell = n+1$. Let the input space be $\mathcal{X}$ and for any $x \in \mathcal{X}$, let $h_0 \in \mathbb{R}^k$ denote the corresponding feature vector. 

\begin{figure}[!t]
\centering
\includegraphics[width=\linewidth]{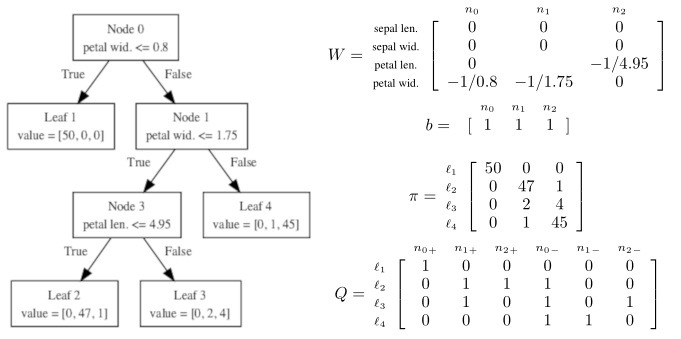}
\caption{Left: Iris Decision Tree by Scikit-Learn, Right: Corresponding Parameters for our Neural Network. Changing the softmax function to a deterministic routing will yield precisely the same result as the Scikit-Learn decision tree.}
\label{fig_tree}
\end{figure}

\begin{figure}[!t]
\centering
\includegraphics[width=3.5in]{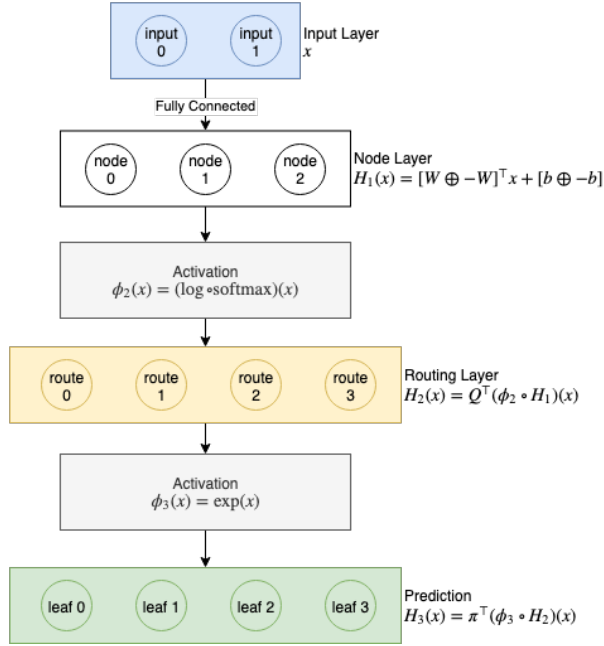}
\caption{Decision Tree as a three layer Neural Network. The Neural Network has two trainable layers: the decision tree nodes, and the leaf nodes.}
\label{fig_nnet}
\end{figure}

The first layer is decision node layer. This is defined by trainable parameters $\Theta = \{ W, b \}$, with $W \in \mathbb{R}^{k \times n}$ and $b \in \mathbb{R}^{n}$. Define $\tilde{W} = [W \oplus -W]$ and $\tilde{b} = [b \oplus -b]$, which represent the positive and negative routes of each node. Then the output of the first layer is $H_1(x) = \tilde{W}^\top x + \tilde{b}$. This is interpreted as the linear decision boundary which dictates how each node is to be routed.

The next is the probability routing layer, which are all untrainable, and are a predetermined binary matrix $Q \in \mathbb{R}^{2n \times (n+1)}$. This matrix is constructed to define an explicit form for routing within a decision tree. We observe that routes in a decision tree are fixed and pre-determined. We introduce a routing matrix $Q$ which is a binary matrix which describes the relationship between the nodes and the leaves. If there are $n$ nodes and $\ell$ leaves, then $Q \in \{0, 1\}^{(\ell \times 2n)}$, where the rows of $Q$ represents the presence of each binary decision of the $n$ nodes for the corresponding leaf $\ell$. We define the activation function to be $\phi_2(x) = (\log \circ \text{ softmax}) (x)$. Then the output of the second layer is $H_2(x) = Q^\top (\phi_2 \circ H_1)(x)$. As $\log(x)$ is 1-Lipschitz bounded function in the domain $(0, 1)$ and the range of $\text{softmax} \in (0, 1)$, then by extension, $\phi_2(x)$ is a 1-Lipschitz bounded function for $x \in \mathbb{R}$. As $Q$ is a binary matrix, then the output of $H_2(x)$ must also be in the range $(-\infty, 0)$. 

The final output layer is the leaf layer, this is a fully connected layer to the previous layer, which is defined by parameter $\pi \in \mathbb{R}^{n+1}$, which represents the number of leaves. The activation function is defined to be $\phi_3(x) = \exp(x)$. The the output of the last layer is defined to be $H_3(x) = \pi^\top (\phi_3 \circ H_2(x))$. Since $H_2(x)$ has range $(-\infty, 0)$, then $\phi_3(x)$ is a 1-Lipschitz bounded function as $\exp(x)$ is 1-Lipschitz bounded in the domain $(-\infty, 0)$. As each activation function is 1-Lipschitz functions, then our decision tree neural network belongs to the same family of artificial neural networks defined by \cite{cortes17a}, and thus our decision trees have the corresponding generalisation error bounds related to AdaNet. 

The formulation of these equations and their parameters is shown in figure \ref{fig_tree} which demonstrates how a decision tree trained in Python Scikit-Learn can have its parameters be converted to a neural decision tree, and figure \ref{fig_nnet} demonstrates the formulation of the three layer network which constructs this decision tree.

\subsubsection{Extending Neural Decision Trees to ResNet}

For our Neural Decison Tree ResNet, in order to ensure that the feature representation is invariant to the number of leaves in the tree, we add a linear projection to ensure that the shortcut connection match the dimensions, as suggested in the original ResNet implementation \cite{He2016DeepRL}.

In this way, we have demonstrated construction of our variations of residual modules retain generalization bounds proved by \cite{cortes17a} and retain true out-of-core online \emph{boosted} learning, compared with other existing algorithms such as BoostResNet\cite{HuangALS17}.


\section{Experiments}\label{experiments}

Below, we perform experiments on two different ResNet architectures. 

First, we examine the ResNet convolution network variant\cite{He2016DeepRL}, with and without the addition of trainable shrinkage parameter. Both models are assessed over street view house numbers SVHN \cite{svhn}, and CIFAR-10 \cite{cifar10} benchmark datasets. 

Second, we examine the efficacy of creating boosted decision tree models in ResNet framework. Our approach was compared against other neural decision tree ensemble models and offline models including \emph{Deep Neural Decision Forests} \cite{dndf}, neural decision trees ensembled via \emph{AdaNet} \cite{cortes17a}, and off-the-shelf algorithms (gradient boosting decision tree/random forest) using \emph{LightGBM} \cite{lightgbm}. All models were assess using UCI datasets which are detailed in Section B of the appendix. 

In both scenarios, the datasets were divided using a $70:30$ split into a training and test dataset respectively. 

\subsection{Convolution Network ResNet}

In both the CIFAR-10 and SVHN datasets we fit the same 20-layer ResNet. This ResNet consists of one $3 \times 3$ convolution, followed by stacks of $18$ layers with $3 \times 3$ convolutions of the feature maps sizes of $\{32, 16, 8\}$ respectively, with $6$ layers for each feature map size. The number of filters are $\{16, 32, 64\}$. The subsampling is performed by convolutions with a stride of 2 and the network ends with global average pooling, a $10$-way fully connected layer and softmax. The implementation is taken directly from the  \href{https://github.com/keras-team/keras/blob/master/examples/cifar10_resnet.py}{Keras CIFAR-10 ResNet} sample code. The model was run without image augmentation, and with a batch size of $32$ for $200$ epochs. To compare the original ResNet, we augment the ResNet model by adding a trainable shrinkage parameter as described in Section \ref{remark} (\emph{ResNet-Shrinkage}), and our augmented ResNet model with both shrinkage parameter and shared linear layer (\emph{ResNet-Shared}).


\begin{table}[t]
\caption{Accuracies of SVHN Task. All trained with same number of iterations (200 epoch, with learning schedule as defined in original ResNet model.)}
\vskip 0.15in
\begin{center}
\begin{small}
\begin{sc}
\begin{tabular}{lrr}
\toprule
Model & Train & Test\\
\midrule
ResNet & 0.93886 & 0.94630  \\ 
\hdashline
ResNet (Shrinkage)  & \textbf{0.93917} & \textbf{0.94852}   \\ 
ResNet (Shared) & 0.93689 & 0.94626 \\
\bottomrule
\end{tabular}
\end{sc}
\end{small}
\end{center}
\vskip -0.1in
\end{table}

\begin{table}[t]
\caption{Accuracies of CIFAR-10 Task. All trained with same number of iterations (200 epoch, with learning schedule as defined in original ResNet model.)}
\vskip 0.15in
\begin{center}
\begin{small}
\begin{sc}
\begin{tabular}{lrr}
\toprule
Model & Train & Test\\
\midrule
ResNet & 0.98412 & 0.91530  \\ 
\hdashline
ResNet (Shrinkage)  & \textbf{0.98504} & \textbf{0.91870}   \\ 
ResNet (Shared) & 0.94496 & 0.88570 \\
\bottomrule
\end{tabular}
\end{sc}
\end{small}
\end{center}
\vskip -0.1in
\end{table}

We find that the model with shrinkage only has marginally higher accuracy than the vanilla ResNet-20 implementation in both datasets. For the \emph{ResNet-Shared} model, it is comparable to the SVHN task, however falls short in the CIFAR-10 task. In general, adding shrinkage does not impact performance of ResNet models and in certain cases, it improves the performance.


\subsection{Neural Decision Tree ResNet}

The next experiment conducted was to address whether ResNet could be used to boost a variety of models with different residual module representations. We compared our decision tree in ResNet (ResNet-DT), and ResNet with shared linear classifier layer (ResNet-DT Shared) with \emph{Deep Neural Decision Forests} \cite{dndf} (DNDF), neural decision trees ensembled via \emph{AdaNet} \cite{cortes17a} (AdaNet-DT), and off-the-shelf algorithms (gradient boosting decision tree/random forest) using \emph{LightGBM} \cite{lightgbm} which we denote as \emph{LightGBDT}, \emph{LightRF} respectively. 

For \emph{ResNet-DT}, \emph{ResNet-DT Shared}, \emph{DNDF}, \emph{LightGBM} and \emph{LightRF}, all models used an ensemble of 15 trees with a maximum depth of 5 (i.e. 32 nodes). For each of these models, they were run for 200 epoch. 

For \emph{AdaNet-DT}, the candidate sub-networks used are decision trees identical to implementation in \emph{DNDF}. This means that at every iteration, a candidate neural decision tree was either added or discarded with no change to the ensemble. The complexity measure function $r$ was defined to be $r(h) = \sqrt{d(h)}$ where $d$ is the number of hidden layers (i.e. number of nodes) in the decision tree \cite{golowich18a}. For \emph{AdaNet-DT}, the algorithm started with $1$ tree, and was run $14$ times with $20$ epoch per iteration, allowing AdaNet to build up to $15$ trees. Once the final neural network structure was chosen, it was run for another 200 epoch and used for comparison with the other models. 

 To assess the efficacy, we used a variety of datasets from the UCI repository. Full results for the training and test data sets are provided in section B of the appendix.

\begin{figure}[ht]

\begin{center}
\centerline{\includegraphics[width=0.65\linewidth]{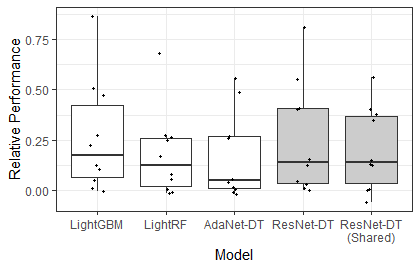}}
\caption{Boxplot of Relative Performance with Deep Neural Decision Forest Model as Baseline on train dataset. High values indicate better performance.}\label{fig:train}
\end{center}

\end{figure}

\begin{figure}[ht]

\begin{center}
\centerline{\includegraphics[width=0.65\linewidth]{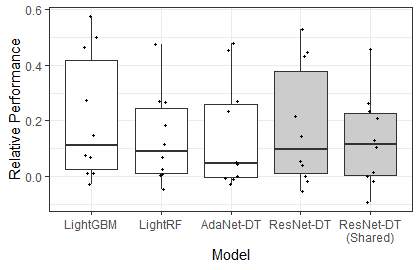}}
\caption{Boxplot of Relative Performance with Deep Neural Decision Forest Model as Baseline on test dataset.  High values indicate better performance.}\label{fig:test}
\end{center}

\end{figure}

\begin{table}[t] \label{tbl:trainuci}
\caption{Mean Improvement compared with Deep Neural Decision Forest Model and Mean Reciprocal Rank on train datasets.}

\begin{center}
\begin{small}
\begin{sc}
\begin{tabular}{lcc}
\toprule
            & Mean  & Mean  \\
            & Improv. & Reciprocal Rank \\
\midrule
LightGBM     & \textbf{26.140\%} & \textit{0.545} \\
LightRF      & 17.414\% & 0.2683 \\
AdaNet-DT    & 16.519\% & 0.345 \\
\hdashline
ResNet-DT    & \textit{25.360\%} & \textbf{0.5783} \\
ResNet-DT (Shared)   & 20.306\% & \textit{0.545} \\
\bottomrule
\end{tabular}
\end{sc}
\end{small}
\end{center}
\end{table}

\begin{table}[t]
\caption{Mean Improvement compared with Deep Neural Decision Forest Model and Mean Reciprocal Rank on test datasets.}
\label{tbl:testuci}

\begin{center}
\begin{small}
\begin{sc}
\begin{tabular}{lcc}
\toprule
            & Mean & Mean  \\
            & Improv. & Reciprocal Rank \\
\midrule
LightGBM     & \textbf{20.904\%} & \textbf{0.67} \\
LightRF      & 13.665\% & 0.3367 \\
AdaNet-DT    & 14.771\% & 0.315 \\
\hdashline
ResNet-DT    & \textit{17.892\%} & \textit{0.5117}  \\
ResNet-DT (Shared)    & 12.949\% & 0.4067 \\
\bottomrule
\end{tabular}
\end{sc}
\end{small}
\end{center}

\end{table}

In order to construct a baseline for all models to be comparable, the results presented are on the average and median error improvement compared with \emph{DNDF} models, as they were the worse performing model based on these benchmarks. From the results in Table \ref{tbl:testuci}, \emph{LightGBM} performed the best with the best average improvement on error relative to the baseline DNDF model. What is interesting is that both our \emph{ResNet-DT} model performed second best, beating \emph{LightRF} and \emph{AdaNet-DT} models.It is important to note that our setup for \emph{AdaNet-DT} only allowed a ``bushy'' candidate model, this did not allow \emph{AdaNet-DT} to build deeper layers compared with \emph{ResNet-DT} approach; only allowing it to build a wider and shallow architect through appending additional decision trees. Despite this, the \emph{AdaNet-DT} implementation did outperform the \emph{DNDF} implementation. 


When examining relative improvement, it is important to understand how the values are then distributed. Figures \ref{fig:train} and \ref{fig:test} contain the boxplots of relative performance based on the train and test datasets respectively. From our empirical experiments, it suggests that the difference between the \emph{ResNet-DT} and \emph{ResNet-DT Shared} are around the variance in the results. One interpretation is through joint training, the variability in performance is lowered and may possible provide more stable models. As to whether joint training should be used or not, we believe it should be considered to be a optional parameter that is learned in training time instead. 

In general, it would appear our \emph{ResNet-DT} performance is comparable to \emph{LightGBM} models whilst providing the ability to update the tree ensemble in an online manner and producing non-greedy decision splits. As this approach can be performed in an online or mini-batch manner, it can be used to incrementally update and train over large datasets compared with \emph{LightGBM} models which can only operate in an offline manner.

\begin{figure}[ht]\label{fig:resmodule}
\vskip 0.2in
\begin{center}
\centerline{\includegraphics[width=0.4\linewidth]{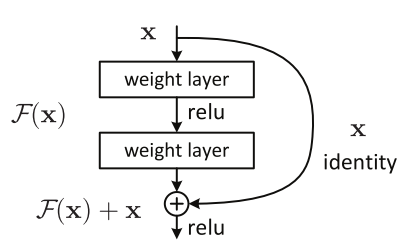}}
\caption{A ResNet-CNN module as defined in \cite{He2016DeepRL}}
\end{center}
\vskip -0.2in
\end{figure}

\subsection{Weak Learning Condition Check}\label{cnn_exp}

We present a summarised proof demonstrating \emph{ResNet-CNN} and \emph{ResNet-DT} satisfy the weak learning condition as stated in Assumption \ref{assum}. The full proof is provided in Section A of the appendix. 

For both cases, it is sufficient to demonstrate that there exists a parameterization such that the residual module $f(x) = x$. Applying this parameterization over the recursive relation $g_{T+1}(x) = \sum_{i=0}^T f_i(g_i(x))$, suggests there exists a parameterization of the residual module $\tilde{g}$ such that $w^\top \tilde{g}_{T+1}(x) \propto w^\top x$. As $w$ is a learnable weight and a linear model, which is a known weak learner \cite{mannor2002existence}, demonstrating that  hypothesis modules created through residual modules are weak learners. 

\textbf{ResNet-CNN: } We will briefly demonstrate that with dense layers in a ResNet setup \cite{He2016DeepRL} can recover the identity. We defer demonstrating convolutional layers scenario to section A of the appendix. 

We will ignore the batch normalization function in ResNet, noting that batch normalization layer with centering value of $0$ and scale of $1$ is a valid parameterization. As such the residual module can be expressed as 

$$f(x) \equiv W^\top \sigma(V^\top x +V_b)  + W_b$$

Where $W$, $V$ are the appropriate weights matrices with $W_b$, $V_b$ being the respective biases and $\sigma$ is ReLu activation. Suppose $W$, $V$ is chosen to be the identity matrix and $W_b$ is chosen to be a matrix containing a single value representing $\min(x, 0)$, and $V_b = -W_b$. Hence there exists a parameterization of \emph{ResNet-CNN} where $f(x) = x$ as required.

\textbf{Remark: } ResNet built under constraints of a linear residual module with only convolution layers and ReLu activations  have been shown to have perfect finite sample expressivity; which is a much stronger condition than recovering only the identity \cite{hardt2016identity}.

\textbf{ResNet-DT: } The weak learning condition can be trivially demonstrated through routing the input $x$ in a deterministic manner to a single leaf with probability $1$. Under this condition the final linear projection layer, project only the target leaf, would result in an identity mapping. This demonstrates a decision tree which routes only to one leaf will have a parameterization $f(x) = x$. This can also interpreted as a ``decision stump'' which is commonly used in boosting applications.

\section{Conclusions and Future Work}

We have demonstrated the equivalence between ResNet and Online Boosting algorithm, and provided a regret bound for ResNet based on the interpretation of residual modules with the linear classifier as weak learners. We have proposed the addition of shrinkage parameters to ResNet, which based on initial results demonstrating it as a promising approach in refining ResNet models. We have also demonstrated a method to remove ``offline'' restriction of BoostResNet of requiring maintaining distribution of all training data weights through extending it to an online gradient boosting algorithm. Together these provide insight into the interpretation of ResNet as well as extensions of residual modules to new and novel feature representations, such as neural decision trees. These representations allow us to create new boosting variations of decision trees. We have additionally demonstrated that this approach is superior to other neural network decision tree ensemble variants and comparable with state-of-the-art offline variations without the drawbacks of offline approaches. In addition we have also provided generalization bounds for our residual module implementations. The insights into the relation between boosting and ResNet could spur other changes to the default ResNet architecture, such as challenging the default size of the step parameter in the \emph{identity skip-connect}. These insights may also change how residual modules are optimized and built, and encourage developments into new residual modules architectures.

\bibliography{main}
\bibliographystyle{IEEEtran}

\clearpage
\appendix

\section*{Appendix}
\subsection{Weak Learners}

To demonstrate that both ResNet-CNN and ResNet-DT we first prove that the existance of the parameterization $f(x) = x$ is a sufficient condition to construct a weak learner. 

\begin{lemma}
If there exists a parameterization $f(x) = x$, then for any $k \in \mathbb{N}^+$, $g_k(x) = 2^{k-1} x$.
\end{lemma}

(By Induction Hypothesis) It is given that $f_0(g_0(x)) = x$ and $g_0(x) = 0$, then using the definition $g_{t+1}(x) = f_t(g_t(x)) + g_t(x)$ and $g_0(x) = 0$, we have $g_1(x) = x+g_0(x) = 2^{1-1} x = x$. 

Assume, for some $k \geq 1$, $g_k(x) = 2^{k-1}x$ holds true, then 
\begin{align*}
g_{k+1}(x) &= f_k(g_k(x)) + g_k(x) \\
&= 2^{k-1}x + 2^{k-1}x \\
&= 2^k x   
\end{align*}

Since both the base case and the inductive step have been performed, then by mathematical induction $g_k(x) = 2^{k-1} x$ for $k \in \mathbb{N}^+$.

\begin{lemma}
If there exists a parameterization $f(x) = x$, then the hypothesis module $\mathbf{w}^\top g_{T+1}(x)$ is a weak learner. 
\end{lemma}

Using Lemma A.1, we can easily see that the hypothesis module $$\mathbf{w}^\top g_{T+1}(x) = 2^T x \propto \mathbf{w}^\top x$$ As $\mathbf{w}$ is a learnable parameter, then the hypothesis module $\mathbf{w}^\top g_{T+1}(x)$ is a weak learner as linear models are weak learners \cite{mannor2002existence}. 

\textbf{ResNet-CNN}

The ResNet-CNN modules generally consist of repeated blocks consisting of convolution-batch normalization-ReLu activation repeated several times (see \verb|resnet50.py| implementation in Keras examples, under \verb|conv_block| and \verb|identity_block|). 

\begin{lemma}
There exists convolution layer and batch normalization weights such that 

$$\tilde{f}(x) = \sigma(\text{BN}(W \cdot x)) = x+c$$

where $W$ is the weights of the convolution layer, $\text{BN}$ is batch normalization function and $\sigma$ is the ReLu function, and $c$ is some constant scalar.
\end{lemma}

As before, we will ignore the batch normalization function in ResNet, noting that batch normalization layer with centering value of $0$ and scale of $1$ is a valid parameterization. Then we construct convolution layer $W$ through choosing only the identity kernel, with bias constructed to be the absolute value of the minimal element in $x$, which we will call $c$. Then 

$$\sigma(\text{BN}(W \cdot x)) = \sigma(W \cdot x) = x+c$$

As all elements in $W\cdot x$ would be greater than $0$, which negates the effect of the ReLu activation function. 

\begin{lemma}

We define the residual module is the composition of arbritary many $\tilde{f}$, as defined below

$$\tilde{F}_k(x)=\begin{cases}
			\tilde{f}(x), & \text{if $k=1$}\\
            (\tilde{f} \circ \tilde{F}_{k-1})(x), & \text{otherwise}
		 \end{cases}$$

Using this definition $\tilde{F}_k(x) = x + c$ for some constant scalar $c$

\end{lemma}

This can trivially be shown via induction. This holds for $k=1$ by Lemma A.3. Assume it holds for $\tilde{k}$, i.e. $\tilde{F}_{\tilde{k}} (x) = x + c$. Then for $\tilde{k}+1$

\begin{align*}
\tilde{F}_{\tilde{k}+1} (x)   &= (\tilde{f} \circ \tilde{F}_{\tilde{k}})(x)\\
&= \tilde{f} (x+c) \\ 
&= x+\tilde{c}
\end{align*}

For some constant scalar $\tilde{c}$. Since both the base case and the inductive step have been performed, then by mathematical induction $\tilde{F}_k(x) = x + c$ for some constant scalar $c$, which suggests that there exists a parameterization for convolution ResNet modules of any depth which recovers the identity.

Therefore using Lemma A.2 and Lemma A.4 the hypothesis module for convolution ResNet model is a weak learner.

\begin{table*}
\caption{The full results are shown below. All datasets are measured based on accuracy}
\label{fullres}
\centering
\resizebox{\textwidth}{!}{%
\begin{tabular}{|l|ll|ll|ll|ll|ll|ll|}
\hline
\multirow{2}{*}{Dataset} & \multicolumn{2}{l|}{AdaNet-DT} & \multicolumn{2}{l|}{ResNet-DT} & \multicolumn{2}{l|}{ResNet-DT (Shared)} & \multicolumn{2}{l|}{DNDF} & \multicolumn{2}{l|}{LightGBM} & \multicolumn{2}{l|}{LightRF} \\
                         & Train          & Test          & Train          & Test          & Train              & Test               & Train       & Test        & Train         & Test          & Train         & Test         \\ \hline
adult                    & 0.8628         & 0.8533        & 0.8837         & 0.8386        & 0.8057             & 0.7733             & 0.8579      & 0.8538      & 0.8638        & 0.8613        & 0.8624        & 0.8596       \\
covtype                  & 0.7153         & 0.7106        & 0.9689         & 0.8302        & 0.9457             & 0.8246             & 0.6877      & 0.6825      & 0.8396        & 0.7831        & 0.8030        & 0.7598       \\
dna                      & 0.9919         & 0.9288        & 0.9888         & 0.8848        & 0.9830             & 0.9204             & 0.9794      & 0.9361      & 0.9745        & 0.9466        & 0.9632        & 0.9393       \\
glass                    & 0.7133         & 0.5781        & 0.7933         & 0.6719        & 0.7933             & 0.5781             & 0.5667      & 0.4688      & 0.8533        & 0.7031        & 0.7067        & 0.5938       \\
letter                   & 0.8524         & 0.8383        & 0.9934         & 0.9700        & 0.9868             & 0.9590             & 0.8602      & 0.8495      & 0.9481        & 0.9075        & 0.9056        & 0.8693       \\
sat                      & 0.8976         & 0.8695        & 0.8904         & 0.8710        & 0.9628             & 0.9125             & 0.8519      & 0.8275      & 0.9556        & 0.8885        & 0.9204        & 0.8835       \\
shuttle                  & 0.9965         & 0.9964        & 0.7860         & 0.7859        & 0.7860             & 0.7859             & 0.7860      & 0.7859      & 0.9997        & 0.9997        & 0.9992        & 0.9985       \\
mandelon                 & 0.8576         & 0.8478        & 0.9848         & 0.9078        & 0.9881             & 0.8856             & 0.8776      & 0.8722      & 0.9219        & 0.8456        & 0.8700        & 0.8322       \\
soybean                  & 0.9937         & 0.9069        & 0.9916         & 0.8873        & 0.9979             & 0.8922             & 0.6388      & 0.6127      & 0.9415        & 0.8971        & 0.8058        & 0.7255       \\
yeast                    & 0.6064         & 0.5618        & 0.7382         & 0.5910        & 0.5486             & 0.4876             & 0.4071      & 0.3865      & 0.7594        & 0.6090        & 0.6843        & 0.5708       \\ \hline
Number of wins & 1 & 1 & 3 & 3 & 3 & 1 & 0 & 0 & 3 & 5 & 0 & 0 \\
Mean Reciprocal Rank & 0.345 & 0.3150 & 0.5783 & 0.5117 & 0.5450 & 0.4067 & 0.1983 & 0.2283 & 0.5450 & 0.6700 & 0.2683 & 0.3367 \\ \hline
\end{tabular}%
}
\end{table*}

\subsection{Description of Data Sets}
The full results are shown in table \ref{fullres}. All datasets are measured based on accuracy.

The datasets used come from the UCI repository and are listed as follows:

\begin{itemize}
\item{\href{https://archive.ics.uci.edu/ml/datasets/adult}{Adult} \cite{adult_data}} 
\item{\href{https://archive.ics.uci.edu/ml/datasets/covertype}{Covertype} \cite{covertype_data}                 }
\item{\href{http://archive.ics.uci.edu/ml/datasets/Statlog+(Landsat+Satellite)}{Landsat Satellite}              }
\item{\href{https://archive.ics.uci.edu/ml/datasets/glass+identification}{Glass Identification}                 }
\item{\href{http://archive.ics.uci.edu/ml/datasets/madelon}{Mandelon} \cite{madelon_data}                       }
\item{\href{https://archive.ics.uci.edu/ml/datasets/soybean+(small)}{Soybean} \cite{soybean_data}                }
\item{\href{https://archive.ics.uci.edu/ml/datasets/Yeast}{Yeast} \cite{yeast_data}                             }
\item{\href{https://archive.ics.uci.edu/ml/datasets/letter+recognition}{Letter Recognition} \cite{letter_data}  }
\item{\href{https://archive.ics.uci.edu/ml/datasets/Statlog+(Shuttle)}{Shuttle}                                 }
\end{itemize}


\end{document}